\pdfoutput=1

\documentclass[11pt]{article}

\usepackage[]{acl}

\usepackage{times}
\usepackage{latexsym,soul,comment}

\usepackage[T1]{fontenc}

\usepackage[utf8]{inputenc}

\usepackage{microtype}

\usepackage{tikz}

\usepackage{amsmath}
\usepackage{amsfonts}
\usepackage{amssymb}
\usepackage{graphicx}
\usepackage{float}

\usepackage{caption}
\usepackage[justification=centering]{caption}
\usepackage[title]{appendix}
\usepackage{subcaption}
\usepackage{adjustbox}

\title{A Girl Has A Name, And It's ...\thanks{~~ This paper is third in the series. See \cite{mahmood2019girl} and \cite{mahmood2020girl} for the first two papers.}\\ Adversarial Authorship Attribution for Deobfuscation\thanks{Our code and data are available at: \url{https://github.com/reginazhai/Authorship-Deobfuscation}}}

\author{Wanyue Zhai \\
  University of California, Davis \\\And
  Jonathan Rusert \\
  University of Iowa \\\\\AND
  Zubair Shafiq \\
  University of California, Davis \\\And
  Padmini Srinivasan \\
  University of Iowa \\}

\begin{document}
\maketitle

\begin{abstract}
Recent advances in natural language processing have enabled powerful privacy-invasive authorship attribution.
To counter authorship attribution, researchers have proposed a variety of rule-based and learning-based text obfuscation approaches. 
However, existing authorship obfuscation approaches do not consider the adversarial threat model. 
Specifically, they are not evaluated against adversarially trained authorship attributors that are aware of potential obfuscation. 
To fill this gap, we investigate the problem of adversarial authorship attribution for deobfuscation. 
We show that adversarially trained authorship attributors are able to degrade the effectiveness of existing obfuscators from 20-30\% to 5-10\%. 
We also evaluate the effectiveness of adversarial training when the attributor makes incorrect assumptions about whether and which obfuscator was used.
While there is a a clear degradation in attribution accuracy, it is noteworthy that this degradation is still at or above the attribution accuracy of the attributor that is not adversarially trained at all.
Our results underline the need for stronger obfuscation approaches that are resistant to deobfuscation.  
\end{abstract}

\section{Introduction}
\label{sec: introduction}
Recent advances in natural language processing have enabled powerful attribution systems\footnote{https://www.eff.org/deeplinks/2013/06/internet-and-surveillance-UN-makes-the-connection}
that are capable of inferring author identity by analyzing text style alone 
\cite{abbasi2008writeprints,narayanan2012feasibility,overdorf16pets,stolerman2013classify,ruder2016character}.
There have been several recent attempts to attribute the authorship of anonymously published text using such advanced authorship attribution approaches.\footnote{https://www.nbcchicago.com/news/politics/Science-May-Help-Identify-Opinion-Columnist-492649561.html}
This poses a serious threat to privacy-conscious individuals, especially human rights activists and journalists who seek anonymity for safety.


Researchers have started to explore text \textit{obfuscation} as a countermeasure to evade privacy-invasive authorship attribution. 
Anonymouth \cite{mcdonald2012use,brennan2012adversarial} was proposed to identify words or phrases that are most revealing of author identity so that these could be manually changed by users seeking anonymity.
Since it can be challenging for users to manually make such changes, follow up work proposed rule-based text obfuscators that can automatically manipulate certain text features (e.g., spelling or synonym)  \cite{mcdonald2013anonymouth,almishari2014cosn,keswani2016author,karadzhov2017case, dcastrocastro2017case,mansoorizadeh2016author, kacmarcik2006obfuscating, Li-delete-retrieve-NAACL-2018}.  
Since then more sophisticated learning-based text obfuscators have been proposed that automatically manipulate text to evade state-of-the-art authorship attribution approaches \cite{karadzhov2017case,shetty2017a4nt,li2018textbugger,mahmood2019girl,grondahl2020effective}.

In the arms race between authorship attribution and authorship obfuscation, it is important that both attribution and obfuscation consider the adversarial threat model \cite{PAN2018overview}. 
While recent work has focused on developing authorship obfuscators that can evade state-of-the-art authorship attribution approaches, there is little work on developing authorship attribution approaches that can work against state-of-the-art authorship obfuscators. 
Existing authorship attributors are primarily designed for the non-adversarial threat model and only evaluated against non-obfuscated documents. 
Thus, it is not surprising that they can be readily evaded by state-of-the-art authorship obfuscators \cite{karadzhov2017case,shetty2017a4nt,li2018textbugger,mahmood2019girl,grondahl2020effective}. 
To fill this gap, we investigate the problem of authorship deobfuscation where the goal is to develop adversarial authorship attribution approaches that are able to attribute obfuscated documents.
We study the problem of \textit{adversarial authorship attribution} in the following two settings. 
\textit{First,} we develop attributors that filter obfuscated documents using obfuscation/obfuscator detectors and then use an authorship attributor that is adversarially trained on obfuscated documents. 
\textit{Second,} we develop adversarially trained authorship attributors that does not make assumptions about whether and which authorship obfuscator is used.

The results show that our authorship deobfuscation approaches are able to significantly reduce the adverse impact of obfuscation, which results in up to 20-30\% degradation in attribution accuracy.
We find that an authorship attributor that is purpose-built for obfuscated documents is able to improve attribution accuracy to within 5\% as without obfuscation. 
We also find that an adversarially trained authorship attributor is  able to improve attribution accuracy to within 10\% as without obfuscation. 
Additionally, we evaluate the effectiveness of adversarial training when the attributor makes incorrect assumptions about whether and which obfuscator is used. 
We find that these erroneous assumptions degrade accuracy up to 20\%, however, this degradation is the same or smaller than when the attributor is not adversarially trained, which can degrade accuracy up to 32\%.

Our key contributions include:

$\bullet$ investigating the \textbf{novel problem} of adversarial authorship attribution for deobfuscation; 

$\bullet$ \textbf{proposing approaches} for adversarial authorship attribution; and  

$\bullet$ \textbf{evaluating robustness} of existing authorship obfuscators against adversarial attribution.

\vspace{.05in} \noindent \textit{Ethics Statement:}
\textcolor{black}{
We acknowledge that authorship deobfuscation in itself is detrimental to privacy. Our goal is to highlight a major limitation of prior work on authorship obfuscation under the adversarial threat model.  
We expect our work to foster further research into new authorship obfuscation approaches that are resistant to deobfuscation.}

\section{Related Work}
    Authorship attribution is the task of identifying the correct author of a document given a range of possible authors.
    It has been a long-standing topic, and researchers have developed a wide range of solutions to the problem. 
    Earlier researchers focus more on analysis based on writing style features. 
    These include the distribution of word counts and basic Bayesian methods \cite{mosteller1963inference}, different types of writing-style features (lexical, syntactic, structural, and content-specific) \cite{zheng2006framework}, and  authors' choices of synonyms \cite{clark2007algorithm}.
    Other researchers combined machine learning and deep learning methods with stylometric features. 
    \citet{abbasi2008writeprints} combine their rich feature set, ``Writeprints'', with an SVM. 
    \citet{brennan2012adversarial} improve ``Writeprints'' to reduce the computational load required of the feature set.
    Finally, more recent research focuses on fine-tuning pre-trained models since they do not require predefined features sets.
    \citet{ruder2016character} tackle authorship attribution with a CNN, while \citet{howard2018universal} introduce the Universal Language Model Fine-tuning (ULMFiT) which shows strong performance in attribution.

    To the best of our knowledge, prior work lacks approaches for adversarial authorship  deobfuscation.
    Prior work has shown that existing authorship attributors do not perform well against obfuscators. 
    \citet{brennan2012adversarial} present a manual obfuscation experiment which causes large accuracy degradation. 
    Since this obfuscation experiment, much has been done in the area of authorship text obfuscation \cite{rao2000can, brennan2012adversarial, mcdonald2012use, mcdonald2013anonymouth,    karadzhov2017case, castro2017author,    mahmood2019girl, grondahl2020effective, bo2019er}. 
    We focus on state-of-the-art obfuscators, Mutant-X \cite{mahmood2019girl} and DS-PAN \cite{castro2017author} specifically in our research. Other obfuscation methods are as vulnerable to adversarial training which is reinforced in \cite{grondahl2020effective}.

    Our proposed authorship attributor leverages adversarial training to attribute documents regardless of obfuscation.
    First described in \cite{goodfellow2014explaining}, adversarial training uses text produced by an adversary to train a model to be more robust. 
    Adversarial training has seen success in other text domains including strengthening word embeddings \cite{miyato2016adversarial}, better classification in cross-lingual texts \cite{dong2020leveraging}, and attacking classifiers \cite{behjati2019universal}.

\section{Methodology} \label{sect:methods}
In this section, we present our approaches for adversarial authorship attribution for deobfuscation.   

\begin{figure*}[!t]
    \subfloat[Scenario 1: \textcolor{black}{Attacker knows the document is obfuscated and the obfuscator used}]
    {\includegraphics[width=.485\linewidth]{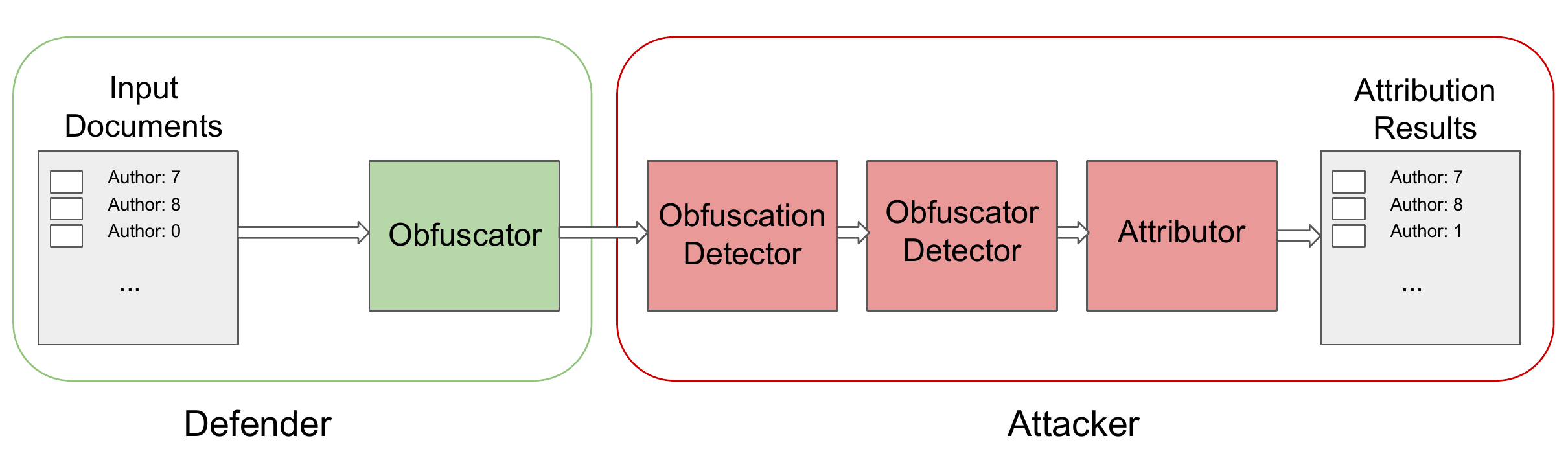} \label{fig:known}}\hfill
    \subfloat[Scenario 2: \textcolor{black}{Attacker only knows the document is obfuscated}]
    {\includegraphics[width=.485\linewidth]{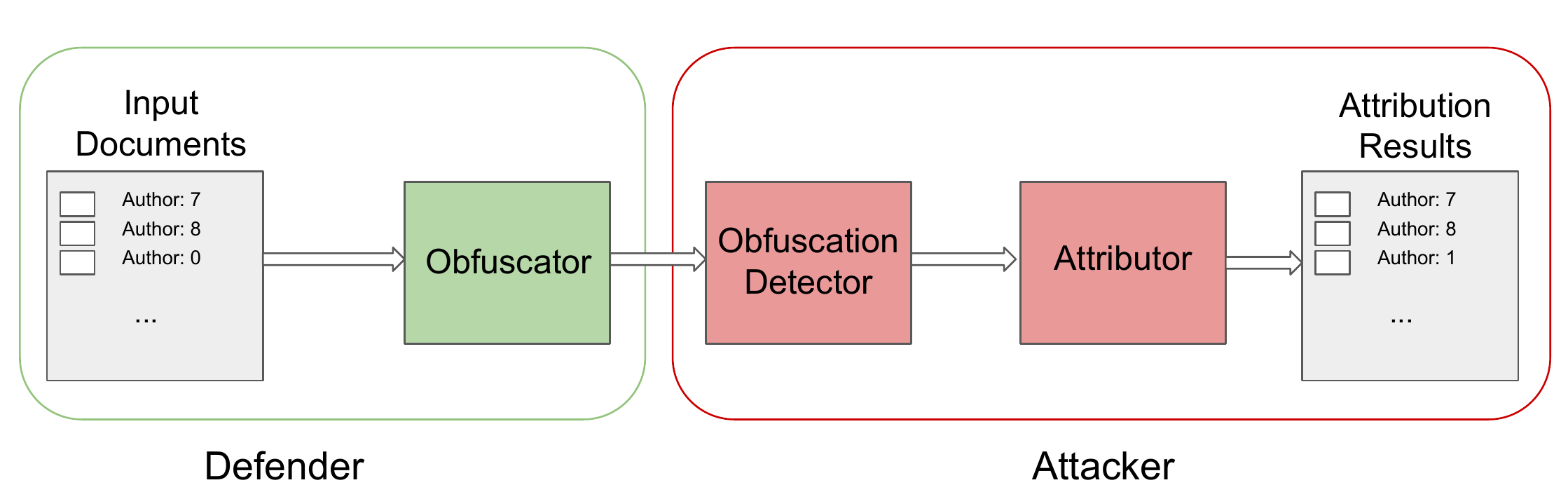} \label{fig:unknown}}\hfill
    \caption{Deobfuscation pipeline using obfuscation and/or obfuscator detectors for adversarial training}
    \label{fig:knownunknown}
    \vspace{- 0.1 in}
\end{figure*}

\subsection{Threat Model} \label{sect:threatmodel}
We start by describing the threat model for the authorship deobfuscation attack.
There is an arms race between an attacker (who desires to identify/attribute the author of a given document) and a defender (an author who desires privacy and therefore uses an obfuscator to protect their identity).
Figure \ref{fig:knownunknown} illustrates the expected workflow between the defender and the attacker. 
The defender uses an obfuscator before publishing the documents and the attacker employs obfuscation and/or obfuscator detector as well as an adversarially trained attributor for deobfuscation.

\vspace{.05in} \noindent \textbf{Defender.} The goal of the defender is to obfuscate a  document so that it cannot be attributed to the author.
The obfuscator takes as input an original document and obfuscates it to produce an obfuscated version that is expected to evade authorship attribution.


\vspace{.05in} \noindent \textbf{Attacker.} The goal of the attacker is to use an attributor trained on documents from multiple authors to identify the author of a given document.
The attacker assumes to know the list of potential authors in the traditional closed-world setting.
We examine two scenarios:
First, as shown in Figure \ref{fig:known}, the attacker assumes to know that the document is obfuscated and also the obfuscator used by the defender. 
In this scenario, the attacker is able to access the documents that are produced by the obfuscator and hence train an attributor for obfuscated documents from the obfuscator.
Second, as shown in Figure \ref{fig:unknown}, the attacker assumes to know that the document is obfuscated and that there is a pool of available obfuscators, of which one is used by the defender. 
Note that the attacker does not know exactly which obfuscator from the pool was used by the defender.
Thus, the attacker trains an attributor for documents that are obfuscated by any one of the pool of available obfuscators.


\subsection{Obfuscation}\label{sect:obfuscators}
We use two state-of-the-art text obfuscators . 

\vspace{.05in} \noindent \textbf{Document Simplification (DS-PAN).} This approach obfuscates documents through rule-based sentence simplification \cite{castro2017author}. 
The transformation rules include lexical transformations, substitutions of contractions or expansions, and eliminations of discourse markers and fragments of text in parenthesis. 
This approach was one of the best performing in the annual PAN competition, a shared CLEF task \cite{potthast2017overview}. 
It was also one of the few approaches that achieves "passable" and even "correct" judgements on the soundness of obfuscated text (i.e., whether the semantics of the original text are preserved) \cite{hagen2017overview}. 
We refer to this approach as DS-PAN.

\vspace{.05in} \noindent \textbf{Mutant-X.}  This approach performs obfuscation using a genetic algorithm based search framework \cite{mahmood2019girl}. 
It makes changes to input text based on the attribution probability and semantics iteratively so that obfuscation improves at each step. 
It is also a fully automated authorship obfuscation approach and outperformed text obfuscation approaches from PAN \cite{potthast2017overview} and has since been used by other text obfuscation approaches \cite{grondahl2020effective}.
There are two versions of Mutant-X: Mutant-X writeprintsRFC, which uses Random Forests along with Writeprints-Static features \cite{brennan2012adversarial}; and Mutant-X embeddingCNN, which uses a Convolutional Neural Network (CNN) classifier with word embeddings.
We use writeprintsRFC version because it achieves better drop in attribution accuracy and semantic preservation as compared to embeddingCNN.



    
\subsection{Deobfuscation}
We describe the design of the authorship attributor and our adversarial training approaches for deobfuscation. 

\noindent
\textbf{Authorship Attributor.} We use writeprintsRFC as the classifier for authorship attribution.
     %
     More specifically, we use the Writeprints-Static feature set \cite{brennan2012adversarial} that includes lexical features on different levels, such as word level (total number of words) and letter level (letter frequency) as well as syntactic features such as the frequency of functional words and parts of speech tags. 
     It is one of the most widely used stylometric feature sets and has consistently achieved high accuracy on different datasets and author sets while maintaining a low computational cost.
     We then use these features to train an ensemble random forest classifier with 50 decision trees. 
     %

\vspace{.05in} \noindent \textbf{Adversarial Training.} 
The basic idea of adversarial training is to include perturbed/obfuscated inputs into the training set to improve the model's resistance towards such adversarially obfuscated inputs \cite{goodfellow2014explaining}. 
It has been widely used in various domains including text classification.
In our case, obfuscated texts are texts that vary slightly from the original texts and these serve as adversarial examples. 
We examine how using these adversarial examples as training data influences the attributor's performance and whether it adds resilience  against obfuscation. 
Based on our two scenarios described in Section \ref{sect:threatmodel} \textcolor{black}{and shown in Figure \ref{fig:knownunknown}}, we propose two ways of adversarial training. 
For both cases, original texts from the list of possible authors are selected and prepared for obfuscation. 
For scenario 1, we train the attributor using documents obfuscated by a known obfuscator. 
For scenario 2, since the attacker does not assume to know the specific obfuscator used by the defender, we train the attributor using documents obfuscated by the pool of available obfuscators.

\section{Experimental Setup}
We describe the dataset, evaluation metrics, and experimental design to assess the effectiveness of our adversarial authorship attribution approaches for deobfuscation.

\vspace{.05in} \noindent \textbf{Dataset.}
Following previous research \cite{mahmood2019girl}, we examine a publicly available dataset for evaluation of our methodology. The Blog Authorship Corpus \cite{schler2006effects} contains over 600,000 blog posts from blogger.com. 
These posts span 19,320 unique authors. 
\textcolor{black}{
Previous research \cite{narayanan2012feasibility} found that authorship attribution gets harder when more authors are included.
Based on the author selection in \cite{mahmood2019girl}, we select a subset of 15 each with 100 documents (compared to their 5 and 10 authors) for a more precised evaluation.} 
These 1500 documents are divided into 80-20\% split for training and testing, respectively. 
{Specifically, 80 documents from each author are used in the training set while the rest 20 documents are used in the test set.}

\begin{figure}[t]
        \centering
        \includegraphics[width=0.45\textwidth]{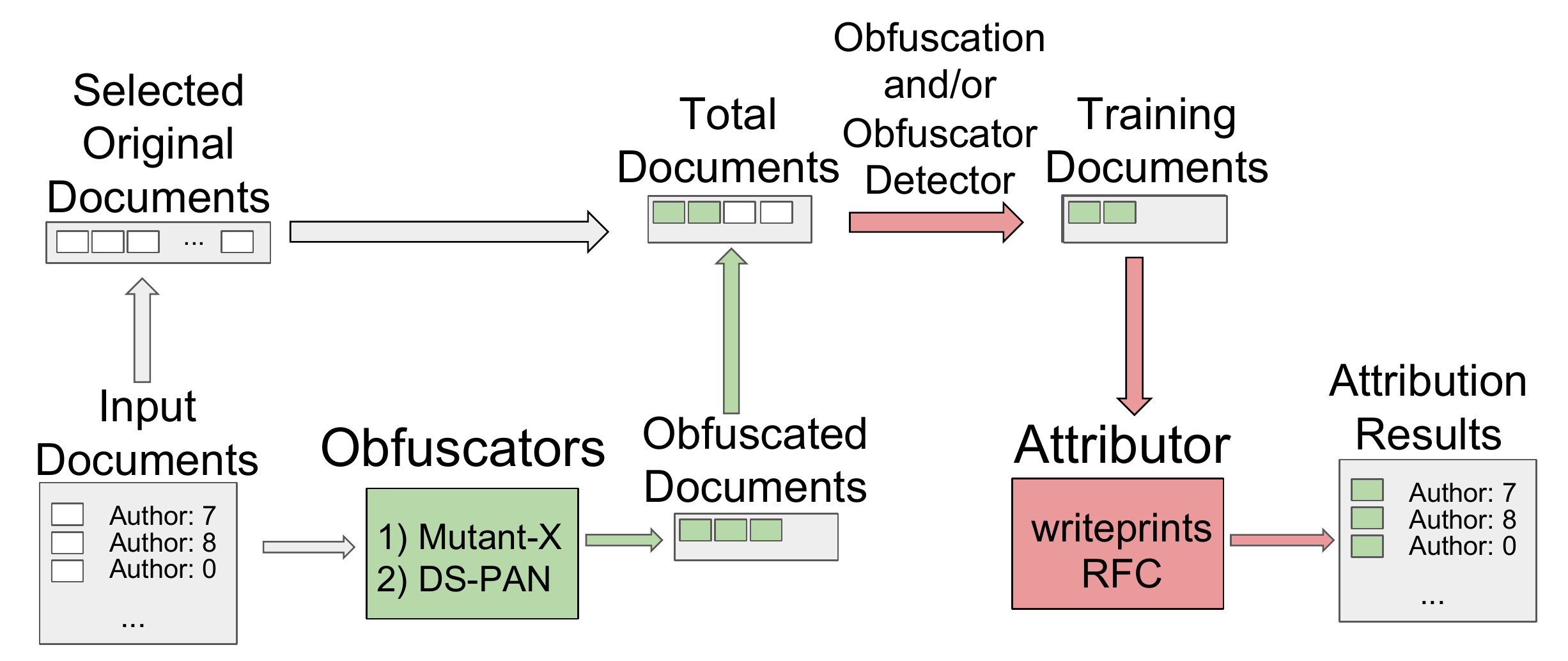}
        \caption{Generalized deobfuscation training process using adversarial training}
        \label{fig:DetailedPi}
        \vspace{- .2 in}
    \end{figure}

As shown in Figure \ref{fig:DetailedPi}, we train on various combinations of obfuscated documents.
These documents are obfuscated by the obfuscators described in Section \ref{sect:obfuscators}. 
When an attributor-dependent-obfuscator (e.g. Mutant-X \cite{mahmood2019girl}) is used, the attributor will have access to the same training documents used to train the obfuscator. 
Otherwise, the attributor does not assume to have access to the attributor used by the obfuscator. 
To control for training size, when more than 1 obfuscator is used, we sample equal amounts of documents from each set of obfuscated documents. 
For example, if we train against 2 obfuscators, then 600 documents are sampled from each set of respective obfuscated documents to get a training set of size 1200.
\textcolor{black}{
To calibrate the obfuscated texts, we use METEOR score \cite{meteorScoreR} to evaluate the soundness of documents.
The score for Mutant-X ranges from 0.3 to 0.7 (mean=0.46), and the score  for DS-PAN ranges from 0.24 to 0.57 (mean=0.38), which are comparable to previous studies \cite{mahmood2019girl}. }
{An in-depth analysis of the METEOR score is reported in Appendix \ref{sec:QulitativeMETEOR}.}


\vspace{.05in} \noindent \textbf{Metric.}
To quantify attribution performance on the 15-class problem, we calculate the accuracy as: 
\begin{equation}
    \text{accuracy} = \frac{\text{\# of correctly attributed documents}}{\text{total \# of documents}}
\end{equation}

\begin{figure}[H]
        \centering
        \includegraphics[width=0.48\textwidth]{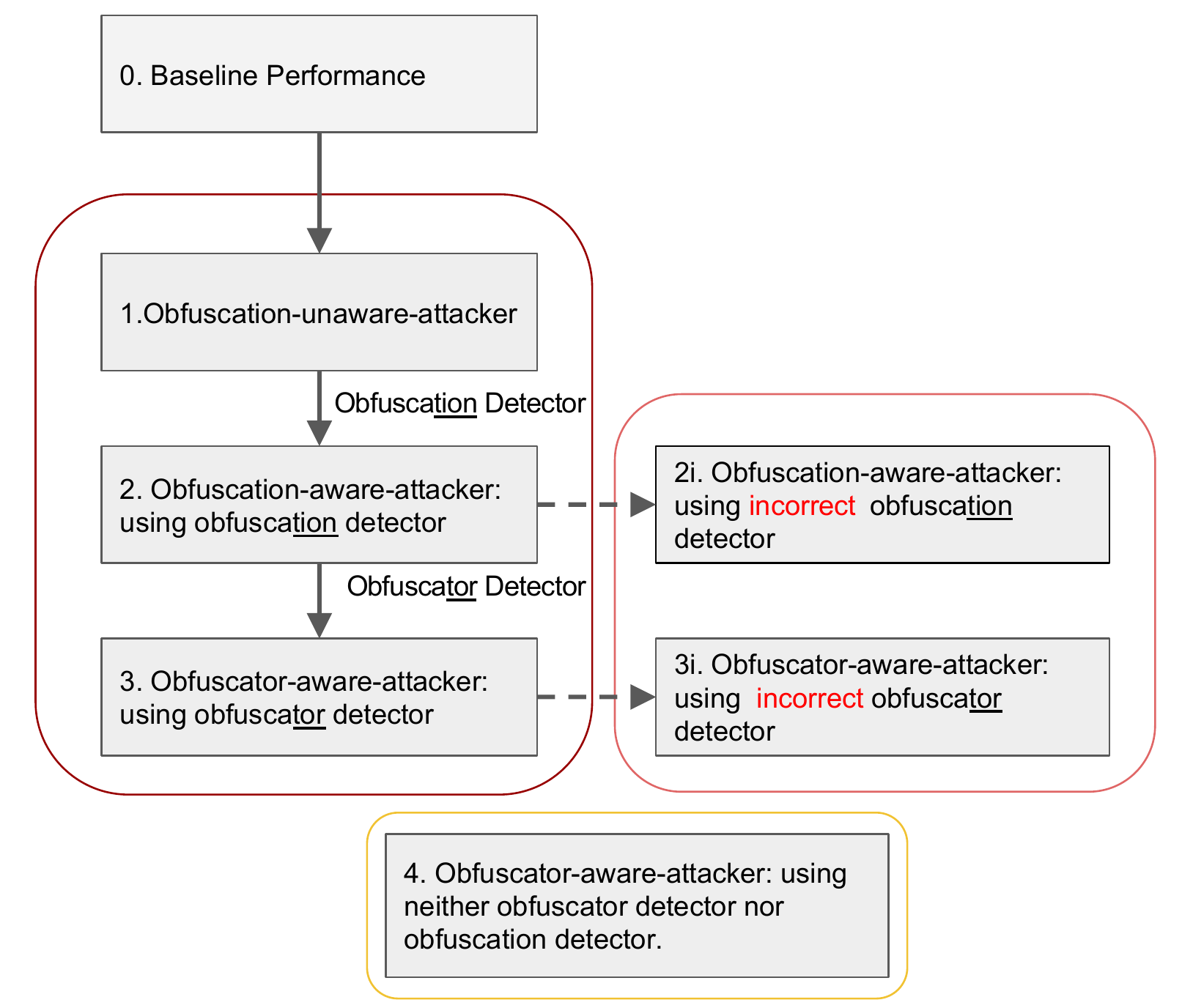}
        \caption{Progression of various attack scenarios}
        \label{fig:result}
        \vspace{- .1 in}
    \end{figure}

\vspace{.05in} \noindent \textbf{Attack Scenarios.}
Figure \ref{fig:result} illustrates the flow of our experimental evaluation under different attack scenarios.

\textbf{0.} \textit{Baseline:}
    For the baseline model, we assume that there is no obfuscation in this world. The attacker is trained on original documents and is deployed on original documents.
    %

\textbf{1.} \textit{Obfuscation-unaware-attacker:}
    The first case we examine is when the defender actively seeks to hide author identity. 
    Thus, the defender gains an advantage by
    obfuscating documents 
    using either Mutant-X or DS-PAN in order to bypass the attacker. 
    %
    The attacker, however, remains unaware of obfuscation and trains the  attributor 
    only on original documents.

\textbf{2.} \textit{Obfuscation-aware-attacker with obfuscation detector:}
    Next, we give knowledge of obfuscation to the attacker by introducing an obfuscation detector into the system. 
    Previous research \cite{mahmood2020girl} shows that  texts generated  by existing obfuscators can be detected as obfuscated  with high accuracy.
    The device for this type of detection is called an obfuscation detector. 
    Hence, in this scenario we ask whether there is any benefit to the attacker if the text is identified as obfuscated before attribution.
    %
    Since the attacker does not know which obfuscator was used by the defender, the attributor is trained on the combination of documents generated from DS-PAN and from Mutant-X.
    The defender is the same as in the previous scenario, i.e., it uses one of the obfuscators to generate documents.

\textbf{3.} \textit{Obfuscator-aware-attacker with obfuscator detector:}
    We give additional information to the attacker.
    Previous research has shown that embedding watermarks and fingerprints \textcolor{black}{that show the identity of the model} into deep neural networks is a valid approach to protect the rights of shared trained models \cite{uchida2017embedding, zhang2018protecting}.
    Hence, it is reasonable to assume that there will be methods in the future to identify the watermarks for specific deep neural networks.
    Here, we propose the concept of obfuscator detector, which can detect the specific obfuscator used.
    %
    In this case, the attacker attributor is trained always on the documents generated by the same obfuscator as the defender: either documents generated from DS-PAN or from Mutant-X.
    %

\textbf{2i.} \textit{Obfuscation-aware-attacker with incorrect obfuscation detector:}
    Here we ask the question: what happens in scenario 2 if the obfuscation detector makes errors?
    The specific error addressed is that the detector classifies the text as obfuscated whereas it is actually an original.
    Under this condition, the attacker attributor is still trained on the combination of documents generated from DS-PAN and from Mutant-X.
    But the defender now presents an original document.

\textbf{3i.} \textit{Obfuscator-aware-attacker with incorrect obfuscator detector:}
    When the obfuscator detector classifies incorrectly, it assumes that the defender uses a specific obfuscator when it actually uses a different one.
    The attacker attributor is trained on the documents generated by one of the obfuscators: either documents generated from DS-PAN or from Mutant-X.
    %
    However, the defender uses a different obfuscator than the attacker to generate the documents.

\textbf{4.} \textit{Obfuscator-aware-attacker that does not rely on an obfuscator detector or obfuscation detector:}
    Since the previous processes require the proposed obfuscation and obfuscator detector, it is not efficient. Hence, a simpler, more efficient solution is to train on all the documents at once.
    In this simplified version, the attacker attributor is trained on the combination of original documents, documents generated from DS-PAN, and documents generated from Mutant-X. 
    Since this is the combined condition, the defender may or may not use an obfuscator, and will choose from the two possible obfuscators to generate documents.

\section{Results}
\label{sec: results}
In this section, we present the results following the progression of various attack scenarios shown in Figure \ref{fig:result}.

\subsection{Major Scenarios}
\textbf{0.} \textit{Baseline:}
    The original authorship attributor has an accuracy of 76.7\% when trained on the original documents and tested on original documents. 
    The attribution accuracy should be higher than 6.7\%, which is when we attribute the 15 authors randomly, to be considered effective.

\textbf{1.} \textit{Obfuscation-unaware-attacker:}
    Based on the first row of Table \ref{tab:shortertable}, the result shows that the attribution accuracy drops from 76.7\% to 50.7\% with a decrease of 26\% when tested on DS-PAN obfuscated documents, while the accuracy for testing on Mutant-X obfuscated documents drops from 76.7\% to 44.3\%. 
    The average drop in accuracy is from 76.7\% to 47.5\%, which is 29.2\%.
    Based on the results, we know that as shown by previous works \cite{karadzhov2017case, mahmood2019girl} \textcolor{black}{on the performance of the obfuscators}, DS-PAN and Mutant-X obfuscators can successfully decrease the attribution accuracy of original attributor.

\textbf{2.} \textit{Obfuscation-aware-attacker with obfuscation detector:}
    The second row of Table \ref{tab:shortertable} shows that attribution accuracy increases by 13.2\% from 50.7\% to 63.9\% when tested on DS-PAN documents, and increases by 24.7\% from 44.3\% to 69\% when tested on Mutant-X documents. 
    The average accuracy is 66.4\%, which increases from the previous 47.5\% by about 19\%.
    While the performance is still not comparable to the baseline results, the increase in accuracy from the previous scenario is significant, which suggests that the obfuscation detector would benefit the performance against obfuscated documents.

\textbf{3.} \textit{Obfuscator-aware-attacker with obfuscator detector:}
    As shown on the third and forth row of Table \ref{tab:shortertable}, when trained only on DS-PAN documents, the accuracy tested on DS-PAN is 68.6\%, with an increase of 17.9\% from the first scenario; when trained only on Mutant-X documents, the accuracy tested on Mutant-X is 75.7\%, with an increase of 31.4\%. 
    The average test accuracy is 71.1\%, which increases by about 5\% compared to the 66.4\% in the previous case. 
    From the results, we can see that having an obfuscator detector as well as an obfuscation detector is the most beneficial to improve the attribution accuracy from obfuscated texts. 
    
    \begin{table}[h]
    \footnotesize
    \centering
    \begin{tabular}{c||c|c||c}
         Training set & \multicolumn{3}{c}{Test set} \\\hline
           & DS-PAN & MutantX & Average\\\hline\hline
          
         Original  & 50.7  & 44.3 & 47.5 \\\hline
         DS-PAN+MutantX  & 63.9 & 69.0 & \textbf{66.4} \\\hline
         DS-PAN  & \textbf{68.6} & - & -\\\hline
         MutantX  & -  & \textbf{75.7} & -\\\hline
    \end{tabular}
    \caption{Accuracy of original attributor and different adversarially trained attributors tested against different obfuscators}
    \label{tab:shortertable}
    \vspace{- .1 in}
    \end{table}
    
\subsection{Error Conditions}
    Although obfuscation/obfuscator detector are quite accurate, they are not perfect. 
    Hence, we test the success of the attacker  when the obfuscation detector and obfuscator detector are incorrect.
    
\textbf{2i.} \textit{Obfuscation-aware-attacker with incorrect obfuscation detector:}
    Shown on the  first column of row four on Table \ref{tab:adversarialResults}, the attribution accuracy decreases by 8.4\% from the baseline 76.7\% to 68.3\%, but a higher accuracy is maintained than the average of Attack Scenario 2 (66.4\%) 
    The result shows that when the obfuscation detector produces wrong results, performance will be influenced, but still stay at a relatively high level. Thus, having an obfuscation detector is generally good for the attacker with little cost.

\textbf{3i.} \textit{Obfuscator-aware-attacker with incorrect obfuscator detector:}
    From second and third rows of Table \ref{tab:adversarialResults} we see that when the attacker is trained only on DS-PAN documents, the accuracy tested on Mutant-X is 57.3\%, with a drop in performance of 18.4\% when compared to training on only Mutant-X documents (75.7\%).
    When the attacker is trained only on Mutant-X documents, the accuracy tested on DS-PAN is 48.5\%, with a drop in performance of 20.1\% as compared to training on only DS-PAN documents (68.6\%). 
    The average test accuracy is 52.9\%, which is lower than training on the same obfuscator, but higher than the results in 1 of 5.1 (50.7\% and 44.3\%).
    When the obfuscator detector gives incorrect results, the attribution accuracy will not achieve its best performance, but the result is still higher than trained only on original documents. Hence, using obfuscated documents to train always tends to benefit the attribution accuracy. 
    

\subsubsection{Combined Condition}
Here the attacker simply uses originals and obfuscated documents from all available obfuscators for adversarial training of the attributor. 

\textbf{4.} \textit{Obfuscator-aware-attacker that does not rely on an obfuscator detector or obfuscation detector:}

    This result is shown on the last row of Table \ref{tab:adversarialResults}. 
    Attribution accuracy when tested on original documents drops from 76.7\% to 66.3\%, but increases by 10.5\% from 50.7\% to 61.2\% when tested on DS-PAN, and increases by 24.5\% from 44.3\% to 68.8\% when tested on Mutant-X. 
    The average accuracy is  65\%, which increases from the average of the former three, 57.2\%, by about 8\%.
    While the attacker does not know if the document is obfuscated or not, or by which obfuscator, it is still able to achieve a high boost in attribution accuracy by adversarial training. 
    Therefore, although the previous processes can achieve higher performances, training on a combination of these documents could be a valid approach when time and resources are limited.

    \begin{table*}[]
    
    \footnotesize
    \centering
    \adjustbox{max width=\textwidth}{%
    \begin{tabular}{c||c|c|c||c}
    
         Training set & \multicolumn{4}{c}{Test set} \\\hline
          & Original & DS-PAN & MutantX & Average of DS+MX\\\hline\hline
          
         Original & \textbf{76.7} & 50.7  & 44.3 & 47.5 \\\hline
         DS-PAN & 57.3 & \textbf{68.6} & 57.3 & 62.9\\\hline
        
         MutantX & 72.0 & 48.5  & \textbf{75.7} & 62.1\\\hline
         


         DS-PAN + MutantX & 68.3 & 63.9 & 69.0 & \textbf{66.4} \\\hline
         
         DS-PAN + MutantX + Original & 66.3 & 61.2 & 68.8 & 65.0 \\\hline

    \end{tabular}}
    \caption{Accuracy of adversarial training on various combinations of test documents}
    \label{tab:adversarialResults}
    
    \end{table*}

    \begin{figure*}[]
    \centering
    \subfloat[Attack Scenario 1\label{fig:3D_ORIG}]
    {\includegraphics[width=.28\textwidth]{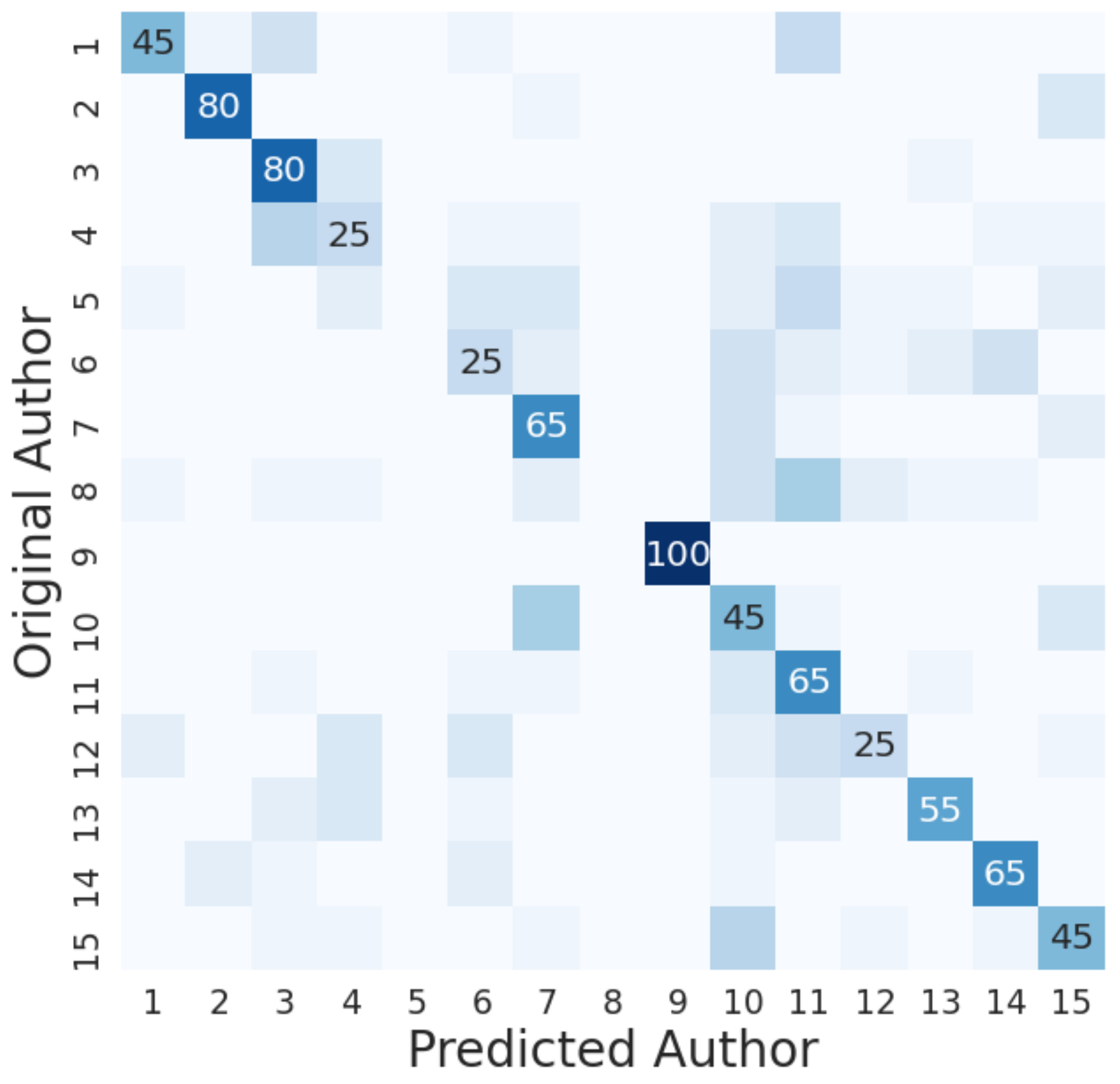}}
    \hspace{0.2cm}
    \subfloat[Attack Scenario 2\label{fig:3D_MIX}]
    {\includegraphics[width=.28\textwidth]{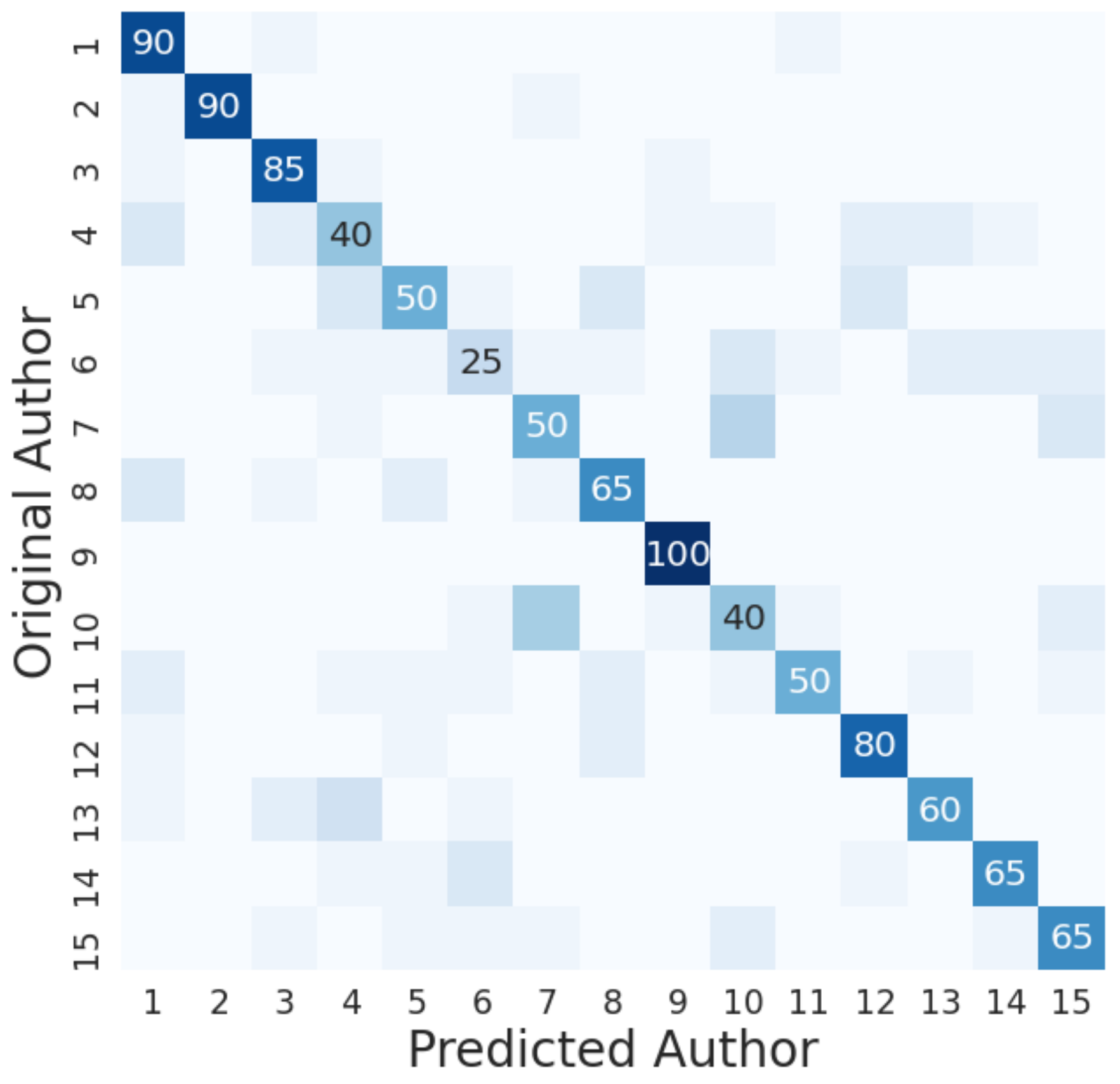}}
    \hspace{0.2cm}
    \subfloat[Attack Scenario 3\label{fig:3D_DS}]
    {\includegraphics[width=.28\textwidth]{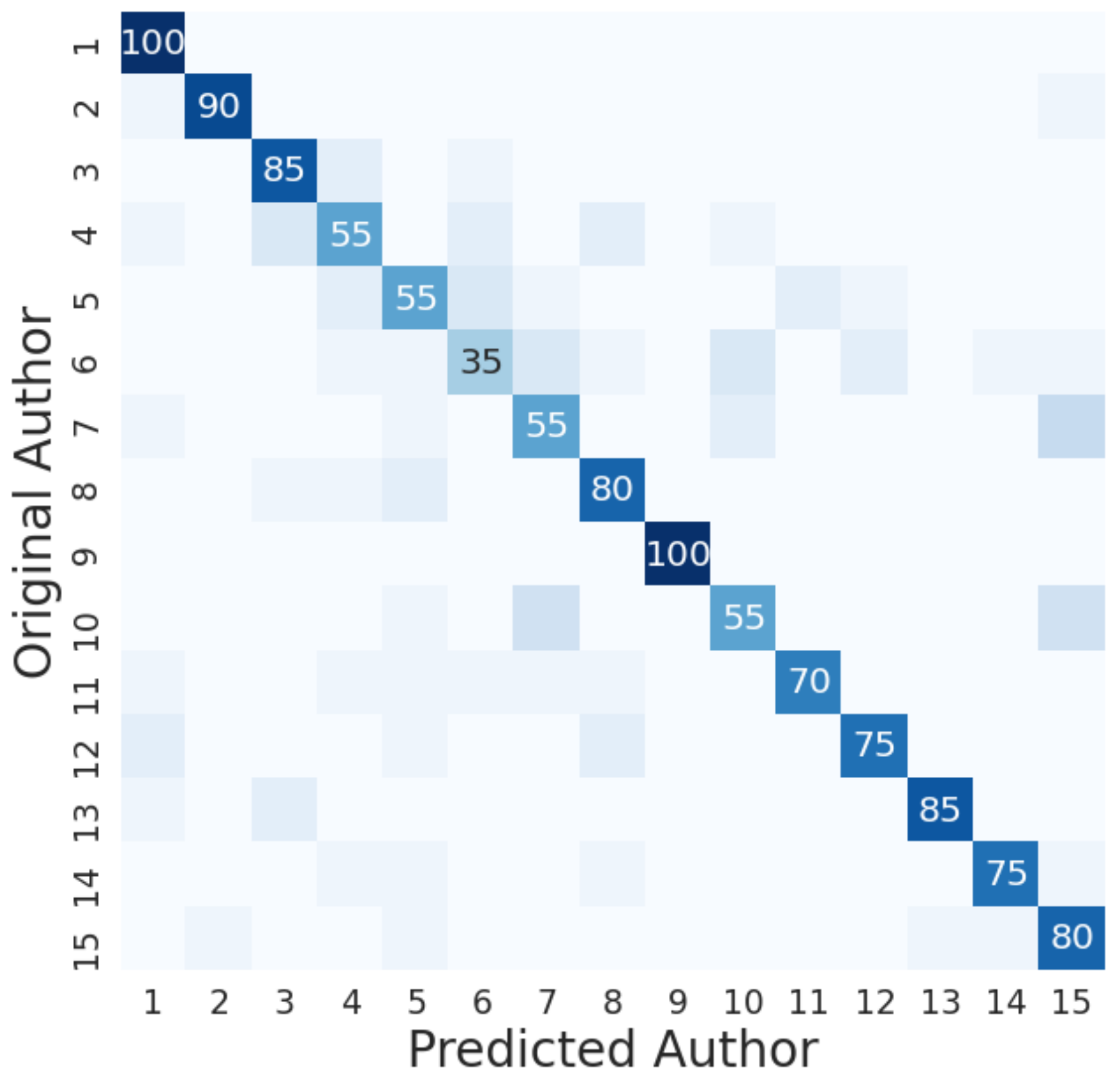}}
    \caption{Confusion matrices of different attack scenarios}
    \label{fig:cm}

    \end{figure*}

    \begin{figure*}[]
    \centering
    \subfloat[Attack Scenario 1\label{fig:3D_ORIG}]
    {\includegraphics[width=.28\textwidth]{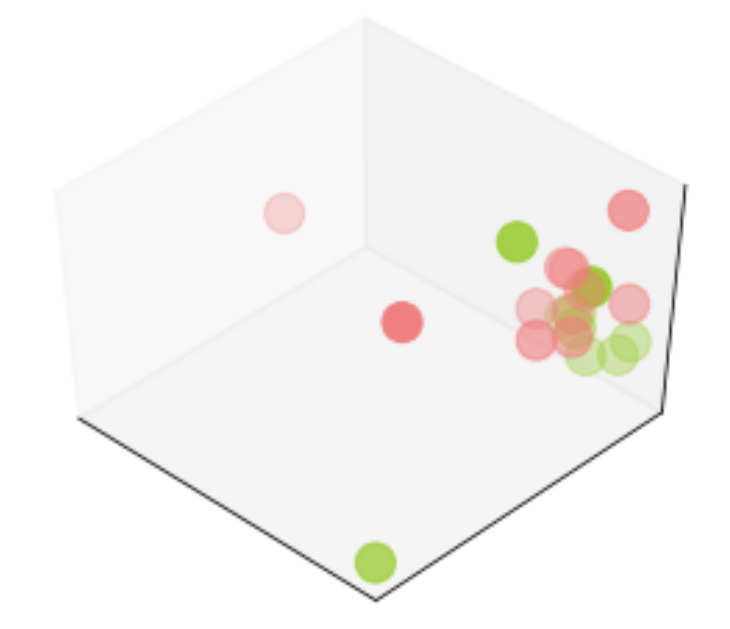}}
    \subfloat[Attack Scenario 2\label{fig:3D_MIX}]
    {\includegraphics[width=.28\textwidth]{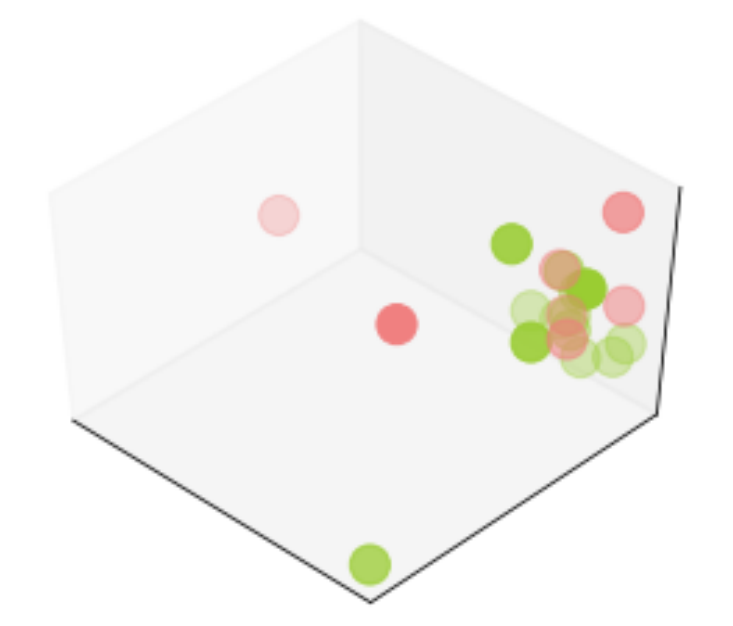}}
    \subfloat[Attack Scenario 3\label{fig:3D_DS}]
    {\includegraphics[width=.28\textwidth]{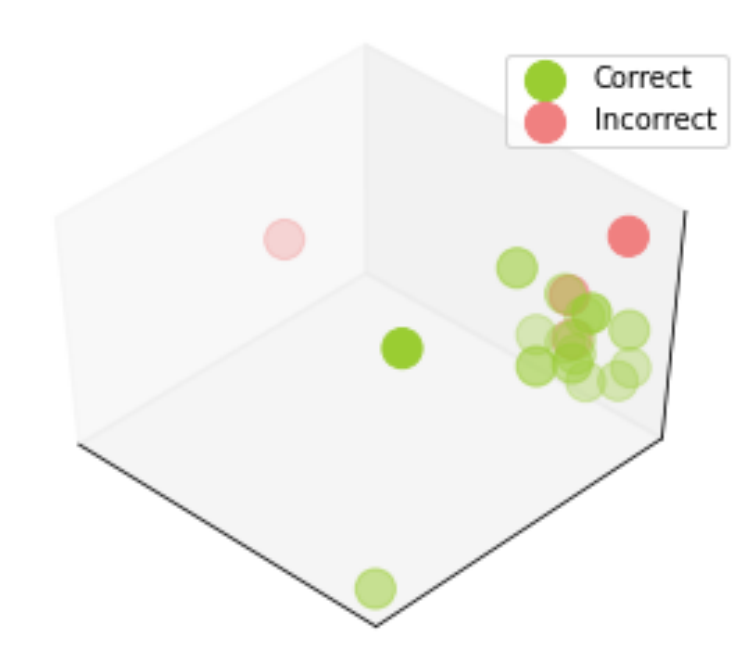}}
    \caption{Attribution performance of Author 15 with PCA under different attack scenarios}
    \label{fig:3D}
    \vspace{- .1 in}
    \hspace*{-0.2cm}
    \end{figure*}

    \begin{figure}[]
        \centering
        \includegraphics[width=0.45\textwidth]{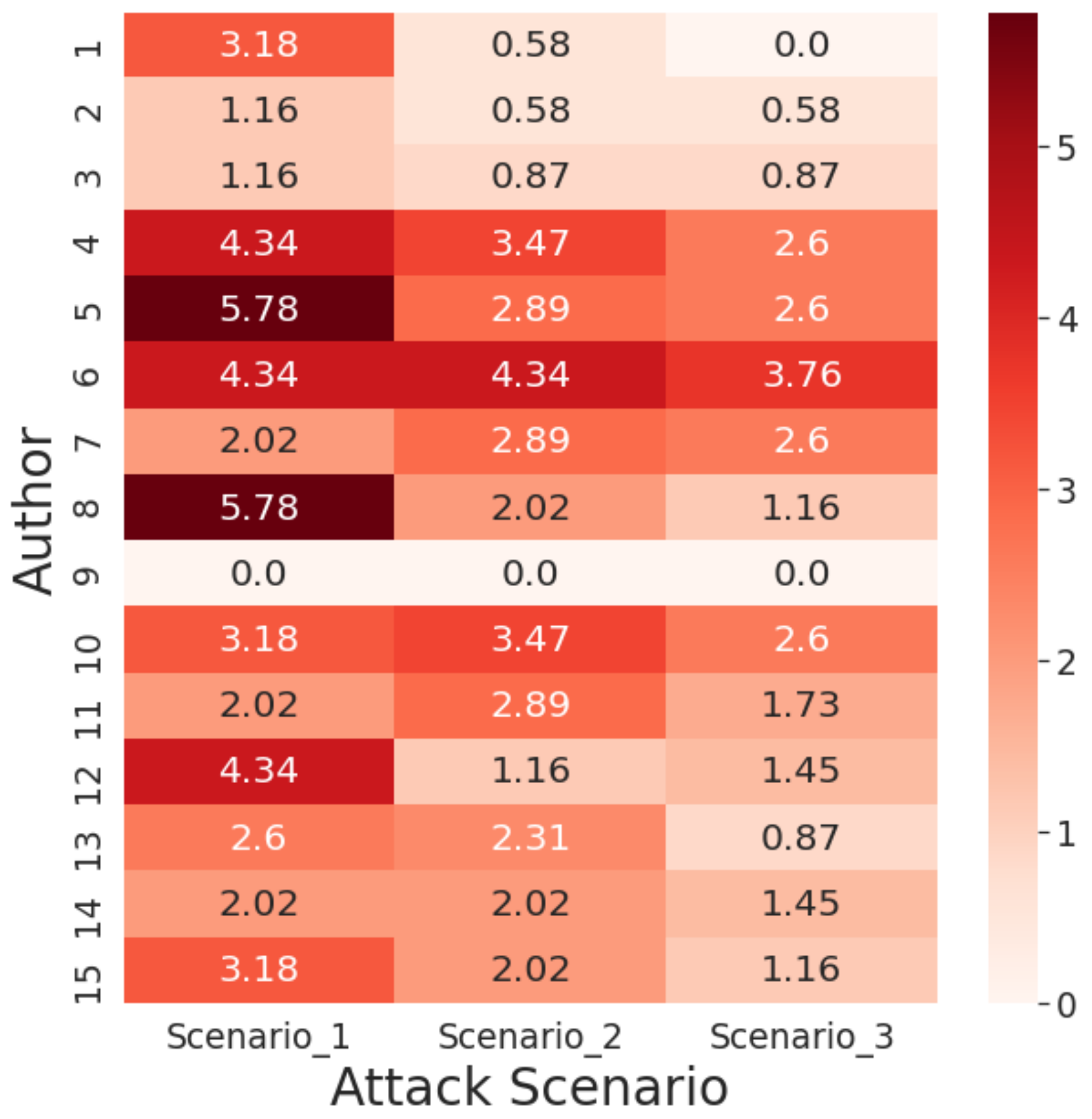}
        \caption{\textcolor{black}{Percentage of misclassified document for each author across attack scenarios}}
        \label{fig:ErrTotal}
    \end{figure}

\section{Discussion}
Next, we look more closely into the results from adversarial training to better understand them. 

\subsection{General Author Analysis}

    %
    Figure \ref{fig:cm} pesents the confusion matrices produced from DS-PAN obfuscated documents tested on Attack Scenario 1, 2 and 3 respectively. 
    Rows represent the Original Authors, while the columns represent the Predicted Authors. 
    The values in the matrices are the percentage of the original documents that are classified as a specific author.

    Moving from scenario 1 to 3, we see an increase in color density and percentage on the diagonal, which signifies the general increase in accuracy when the training documents become more specific. 
    Consistent with above, the color on the non-diagonal areas becoming more transparent also indicates reduction of classification errors.
    At the author level, we observe that almost all of the  authors show  increases in accuracy on the diagonal cells across the three scenarios.
    It shows that adversarial training is effective even on authors with different styles.
    %
    
    %
    Looking more closely at each author, we know that Author 9 is the easiest to classify - performance is always at 100\%. 
    Author 6, on the other hand, is relatively hard to attribute.
    The best performance for Author 6 is only 35\% from the most effective Attack Scenario 3.

    Figure \ref{fig:ErrTotal} presents another view on performance.
    It shows the percentage of errors made for each author out of all the errors in the three scenarios combined (note: the sum of all errors in the figure is 100\%). 
    Thus, the errors made for Author 1 under Scenario 1 is 3.18\% of total errors across the three scenarios.
    We observe that the color is generally darker in Scenario 1, while it gradually lightens in Scenario 2 and then in Scenario 3.
    Again, this indicates the benefit of having more specific training data. 
    %
    Looking more closely within each scenario, we see that the attributor of Attacker Scenario 1 tends to misclassify Authors 5 and 8 the most. 
    But the attributors for Scenario 2 and Scenario 3 learn more effectively for these two authors thereby reducing mistakes.
    For Attack Scenario 3, the most misclassified author is Author 6, where 3.76\% of all errors. 
    But this percentage is still an improvement over the 4.34\% in the previous two scenarios.
    Motivated by the above observations, next we investigate shifts in performance for a specific author.

\subsection{Individual Author Analysis} 
    
    We \textcolor{black}{assign labels to the 15 authors in the dataset and} select Original Author 15 for more detailed analysis. 
    The reason we choose Author 15 is that its accuracy is among the ones that increases the most, from 45\% to 80\%. 
    In order to find out the reasons behind such increase, we perform PCA analysis on all of the DS-PAN documents whose original author is Author 15. 
    We use Writeprints-Static feature set, which has a total of 555 features. 
    In order to preserve the most significant features for attribution, we select the most important 25 features from the original writeprintsRFC and process them through PCA so that we can visualize the features into 3 dimensional graphs.

    As shown in the graphs in Figure \ref{fig:3D}, each dot on the graph represents a document.
    The green ones are the ones that are attributed correctly while the red ones are attributed incorrectly. 
    In Figure \ref{fig:3D_ORIG}, the incorrectly attributed ones are mainly gathered in a cluster. 
    This suggests that the attributor has trouble discriminating the documents that are similar to each other. 
    But as we go from left to right, the documents in the cluster are also gradually attributed correctly. 
    The trend shows that the attributor is getting better at distinguishing between documents that are similar to each other.
    Hence, we can infer that adversarial training improves attribution accuracy by discriminating between the ones that are more similar to each other. 
     
\subsection{Comparing DS-PAN and Mutant-X} 
    %
    %
    In Attack Scenarios 2, 3, and 4, the test sets using DS-PAN for obfuscation yield worse attribution accuracy than those using Mutant-X.
    Our analysis of obfuscated documents showed that DS-PAN makes both a greater number of changes as well as more significant changes as compared to Mutant-X. 
    Thus, we surmise that DS-PAN results in larger degradation in attribution accuracy because the attacker's training set contains text that is less similar to the original text.
    However, the changes made by DS-PAN also have side effect in that they lower the soundness of obfuscated text as reflected by lower METEOR scores.
    The mean METEOR score for DS-PAN is 0.38 as compared to 0.46 for Mutant-X.
    A more detailed analysis of METEOR score and semantic similarity between obfuscated and original texts is reported in Appendix \ref{sec:QulitativeMETEOR}.
    
\subsection{Insights into Adversarial Training} 
    The performance gain of adversarial training comes from a "noisy"  training dataset comprising of obfuscated documents as well as knowledge about the obfuscator. 
    To disentangle these two factors, we compare the accuracy improvements of the second and third rows of Table \ref{tab:adversarialResults} against the Mutant-X obfuscated test documents.
    We note that the improvement in attribution accuracy is 13\% when DS-PAN obfuscated documents are used for training. 
    The improvement in attribution accuracy is further 18\% (31\% overall) when Mutant-X obfuscated documents are used for training. 
    This difference (13\% vs. 18\%) indicates that although having a noisy dataset helps, the knowledge of the specific obfuscator is likely more crucial to improving attribution performance. 
    This trend holds for DS-PAN obfuscated test documents. 
    





\section{Concluding Remarks}
\label{sec: conclusion}
In this work, we explored the novel problem of adversarial authorship attribution for deobfuscation. 
We demonstrate that adversarial training is able to significantly reduce the adverse impact of existing text obfuscators on authorship attribution accuracy. 
We found that an adversarially trained authorship attributor improves attribution accuracy to within 5-10\% as without obfuscation.
While an adversarially trained authorship attributor achieved best accuracy when it is trained using the documents obfuscated by the respective obfuscator, we found that it achieves reasonable accuracy even when it is trained using documents obfuscated by a pool of obfuscators. 
When the adversarially trained attributor makes erroneous assumptions about the obfuscator used to obfuscate documents, we note a degradation in attribution accuracy. 
It is noteworthy, however, that this degradation is still similar or better than the attribution accuracy of the baseline attributor that is not adversarially trained.


%

%


Our results shed light into the future of the ensuing arms race between obfuscators and attributors. 
Most notably, we find that the effectiveness of adversarial training is somewhat limited if the obfuscators continue to employ new and improved methods that are not available to attributors for adversarial training. 
Therefore, it is important to continue development of new and improved text obfuscation approaches that are resistant to deobfuscation \cite{bevendorff2019heuristic,bo2019er,grondahl2020effective,hlavcheva2021language}. 
On the other hand, recent work on understanding and improving transferability of adversarial attacks can inform development of better adversarial attributors that might work well even for unknown obfuscators \cite{tramer2017space,zheng2020efficient,he2021model,Mireshghallah21stylepooking}.

Finally, our experiments were limited to the closed-world setting where the universe of potential authors is assumed to be known by the attributor. 
Further research is needed to investigate whether (and how much) adversarial algorithms are effective in the open-world setting. 
%

%



\bibliographystyle{acl_natbib}
\bibliography{main}

\appendix
\onecolumn



\twocolumn
\section{Qualitative Analysis} 
\label{sec:QulitativeMETEOR}

    We conduct analysis to evaluate the quality of the text. We first evaluate the semantics of the obfuscated text with respect to the original text using METEOR scores. The results show that METEOR scores of obfuscated text are comparable to those reported in prior studies. We also conduct qualitative analysis of the obfuscated text. 

    First, we evaluate the quality of obfuscated documents from the two obfuscators. 
    We use METEOR score to measure the soundness of the obfuscated text in terms of the semantic similarity between the original and the obfuscated text.
    
    \begin{figure}[ht]
        \centering
        \includegraphics[width=0.45\textwidth]{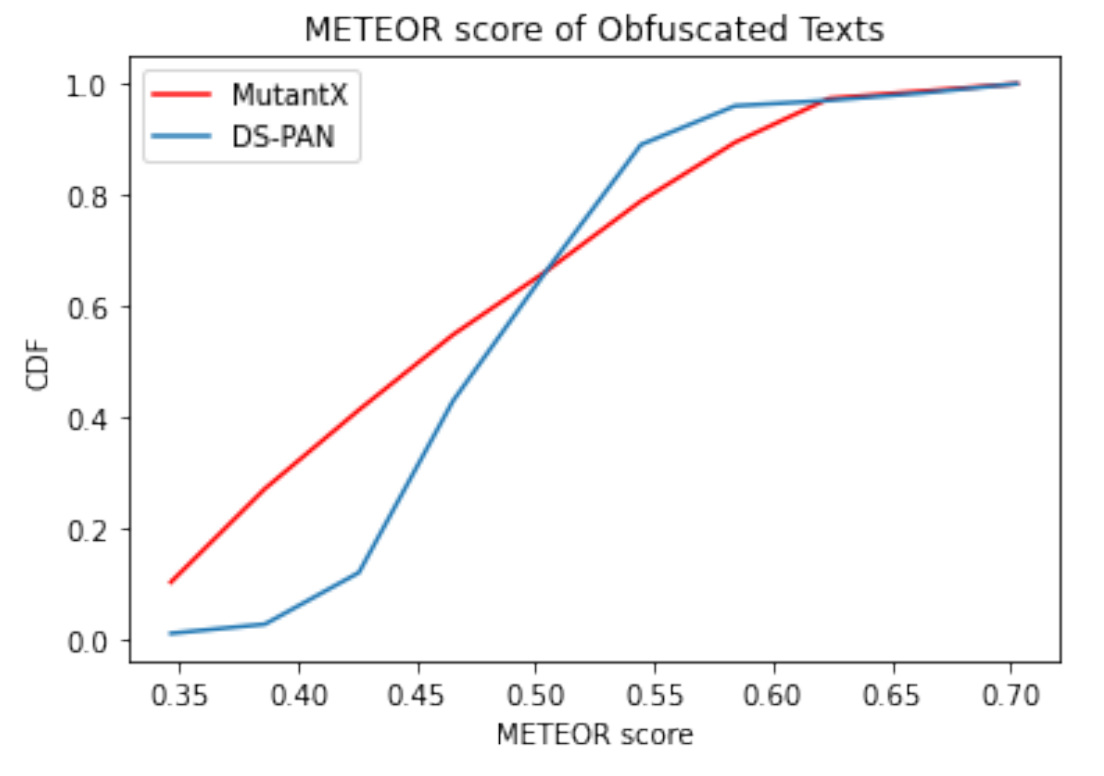}
        \caption{{CDF plot of METEOR score for obfuscated texts}}
        \label{fig:meteor}
    \end{figure}

    \begin{table*}[]
    \footnotesize
    \centering
    \begin{tabular}{|p {0.7cm}|p{3.75cm}||p{3.75cm}|p{3.75cm}|}
        \hline
         Index & Original & DS-PAN & MutantX\\\hline\hline
         1 & I'm not an expert & I'm not An expert & I am non an expert \\ \hline
         2 & What was the first print run? & What was the first print running? & What was the ane print run? \\ \hline
         3 & The New York Times ran a Styles section profile two weeks before publication & The New York Times ran a Styles editor profile two weeks before publication & the new\_york\_times run a styles division profile two calendar\_week before publishing \\ \hline \hline
         
         4 & Cornelius walks in off of the street. & Cornelius walks in off of the sidewalk & Cornelius walks in away of the street. \\ \hline
         5 & We've discovered librarians are very networked and seem to know about everything before it happens & We've found librarians are extremely networked and seem to believe about everything before it happens. & we suffer detect bibliothec are really network and appear to cognize about everything before it happen \\ \hline
         6 & Homework is minimal, but the reading load is daunting. & Homework is minor, but the reading load is daunting. & Prep is minimum, but the read load is daunt \\ \hline \hline
         7 & Some traces of the original layout remain & Some traces of the manifest makeover remain & Some trace of the original layout stay \\ \hline
         8 & Some professors seem happy to have a visitor & Some professors seem happy to become a pilgrim & Some prof appear happy to've a visitor \\ \hline
         9 & He expects interest in the Nancy Pearl doll to be strongest in Seattle, where she is best known. & He expects grateful in the Nancy Pearl mannequin to be strongest in Seattle, where she is best known. & He expect involvement in the nancy\_pearl dolly to be strongest in seattle, where she's well cognize. \\ \hline
         10 & When the sales slot came open a few months later, she applied. & When the sales position came open a few years later, she applied. & When the cut-rate\_sale time\_slot arrive open\_up a few calendar\_month she utilize. \\ \hline
         11 & Professors often mistake her for a student & Professors often mistake her for a campus & Prof frequently err her for a pupil \\ \hline \hline
         
         12 & They may look sleepy, but many used-book stores are thriving. & They may look sleepy, although many used-book stores are mature & they may search sleepy-eyed, but many used-book stores are boom \\ \hline
         13 & The perfumed bear she gave to me lost his scent & The perfumed bobcat she gave to me lost his odor & The perfume bear she render to me lose his aroma \\ \hline
        14 &  I suppose I would have just waited until the morning if I were her. & I reckon I will rest just waited until the afternoon if I were She. & I presuppose i'd suffer precisely wait until the morn if i were her. \\ \hline

    \end{tabular}
    \caption{Sentences from test document showing the result of different obfuscators}
    \label{tab:example}
    \end{table*}

    Figure \ref{fig:meteor} shows the distribution of the METEOR score for Mutant-X and DS-PAN. The plot shows that the METEOR scores for Mutant-X ranges from 0.3 to 0.7 (mean=0.46), and the METEOR score for DS-PAN ranges from 0.24 to 0.57 (mean=0.38). Compared to the previous METEOR score results calculated in \cite{mahmood2019girl}, where the METEOR score for Mutant-X ranges from 0.48 to 0.55 (mean = 0.51), and the METEOR score for other baseline models ranges from 0.32 to 0.46 (mean = 0.38), the two obfuscators used in this work achieve similar results at preserving the semantics of the original texts. 

    Table \ref{tab:example} contains examples from the two obfuscators showing different types of changes. Synonym replacement is common in both systems. Examples of such are (street <-> sidewalk), (student <-> pupil). There are also changes in word form. (run <-> running), (waited <-> wait) preserves the morpheme, but changes the tense of the word. It is also worth noting that DS-PAN tends to change the form of abbreviations, such as (I'm <-> I am) and (to have <-> to've). In general, the transformations make sense to the readers, and preserve most of the original meanings. But there are also cases (like the last row) where the transformations change the content and break the grammar.
    

\end{document}